\documentclass[10pt,twocolumn,letterpaper]{article}

\usepackage[pagenumbers]{cvpr}      

\usepackage{graphicx}
\usepackage{amsmath}
\usepackage{amssymb}
\usepackage{booktabs}

\usepackage{xspace}
\usepackage{dsfont}
\usepackage{multirow}
\usepackage{xcolor}

\usepackage{epstopdf}

\usepackage[pagebackref,breaklinks,colorlinks]{hyperref}

\usepackage[capitalize]{cleveref}
\crefname{section}{Sec.}{Secs.}
\Crefname{section}{Section}{Sections}
\Crefname{table}{Table}{Tables}
\crefname{table}{Tab.}{Tabs.}

\def\cvprPaperID{4791} 
\def\confName{CVPR}
\def\confYear{2022}

\newcommand{\norm}[1]{\left\lVert#1\right\rVert}

\makeatletter
\DeclareRobustCommand\onedot{\futurelet\@let@token\@onedot} 
\def\@onedot{\ifx\@let@token.\else.\null\fi\xspace}

\def\eg{\emph{e.g}\onedot} 
\def\ie{\emph{i.e}\onedot} 
 
 \def\vs{\emph{vs}\onedot}
 
\def\etal{\emph{et al}\onedot}
\makeatother

\begin{document}

\title{Long-tail Recognition via Compositional Knowledge Transfer}

\author{Sarah Parisot\qquad 
Pedro M. Esperan\c{c}a \qquad 
Steven McDonagh \qquad 
Tamas J. Madarasz \qquad \\
Yongxin Yang \qquad
Zhenguo Li  \\ \\
Huawei Noah's Ark Lab
}

\maketitle

\begin{abstract}
In this work, we introduce a novel strategy for long-tail recognition that addresses the tail classes' few-shot problem via training-free knowledge transfer. Our objective is to transfer knowledge acquired from information-rich common classes to semantically similar, and yet data-hungry, rare classes in order to obtain stronger tail class representations. We leverage the fact that class prototypes and learned cosine classifiers provide two different, complementary representations of class cluster centres in feature space, and use an attention mechanism to select and recompose learned classifier features from common classes to obtain higher quality rare class representations. 
Our knowledge transfer process is training free, reducing overfitting risks, and can afford continual extension of classifiers to new classes. 
Experiments show that our approach can achieve significant performance boosts on rare classes while maintaining robust common class performance, outperforming directly comparable state-of-the-art models. 
\end{abstract}

\section{Introduction}
\label{sec:intro} 

Standard classification models rely on the assumption that all classes of interest are equally represented in training datasets.
This strong assumption is rarely valid in practice: most real life datasets exhibit long-tail distributions where a subset of classes comprise a large amount of training data (the so-called common or head classes) and remaining tail (rare) classes only possess a handful of training samples~\cite{OLTR}.
These distributions typically reflect what is observed in the wild, with data corresponding to tail classes being more challenging to learn due to sample rarity, acquisition and collation costs or combinations of these factors. 
This results in trained models that display highly imbalanced accuracies, achieving strong performance on common classes, and poor performance on rare classes.

Long-tail distributions present two main challenges for recognition models. 
Firstly, the imbalanced distribution of class specific data leads to models exceedingly biased towards common classes. 
Numerous strategies have been developed to address this imbalance challenge. Popular strategies aim to rebalance the training process via biased resampling~\cite{balanced_softmax,OLTR,decouple,bbn}, rebalancing loss functions~\cite{balanced_softmax,lade,causalnorm,paco,weightedCE}, or ensembling and routing~\cite{ride,tade,reslt}. 

\begin{figure}[t]
    \centering
    \includegraphics[width=\linewidth]{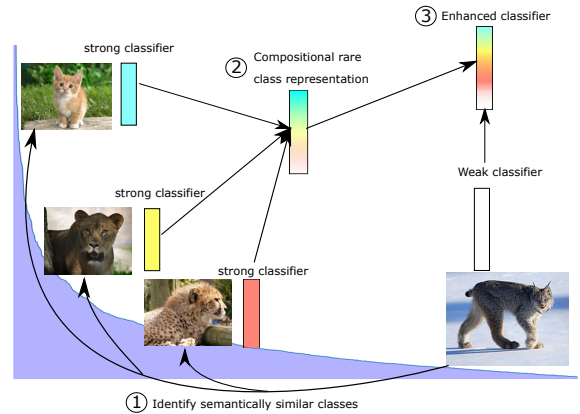}
    \caption{Our approach for long-tail recognition seeks to improve representations for rare classes via knowledge transfer from common classes. We identify semantically similar head classes using class prototypes and transfer reliable learned characteristics from these classes to obtain higher quality rare class representations. }
    \label{fig:teaser}
\end{figure}

While the class imbalance problem has been extensively studied, a second important challenge associated with long-tail distributions has received only limited attention. Rare classes on the distribution tail typically have very limited amounts of data available (\eg $1$ to $20$ training samples per class), which can easily lead to overfitting or memorisation when attempting to learn class specific decision boundaries. This issue is exacerbated by rebalancing strategies, in particular those relying on oversampling data from the tail. 
However approaches dedicated to long-tail recognition very rarely include specific mechanisms to account for the limited tail data. 
A synthetic sample generation process was proposed in \cite{RSG} to improve rare class distributions. A knowledge transfer mechanism was proposed in \cite{OLTR}, aiming to learn better feature representations by transferring knowledge from learnable memory vectors, under the assumption that the model can automatically recognise which images require such transfer. The process, however, failed to consider a transfer mechanism for classifier learning, which is where lack of data has the largest impact~\cite{decouple}.

In contrast, few-shot learning (FSL) has been extensively studied in settings that rely on 
strong assumptions, making direct application of existing solutions to the long-tail problem challenging. Indeed, FSL approaches are typically developed so as to perform well on so-called episode benchmarks \cite{miniim,metadataset}, defined as a small group of sampled classes with a fixed, consistent number of training samples per class. These assumptions typically lead to poor generalisation both with regards to common classes and when a variable number of training samples are available~\cite{OLTR}. 

Class prototypes are a popular concept in FSL. They are defined as class representations computed directly from the data, which is advantageous in limited data regimes due to the difficulty of learning reliable few-shot classifiers. Instead of seeking appropriate decision boundaries, prototypical models seek to learn compact and well separated representations by minimising samples' distance with respect to their  class prototypes \cite{protonets}. This can be achieved by explicit prototypes computation via artificially constructed episode batches \cite{protonets}, or implicitly, using cosine-distance based classifiers \cite{imprintedweights, dynamicFSL}. 
Learning distance based feature representations encourages classes with similar semantic meanings to be close in feature space. This provides an advantageous solution to leverage class similarities for our long-tail task.

In this work, we introduce a novel classification strategy for long-tail recognition that addresses the few-shot problem via a training free knowledge transfer mechanism. Our objective is to transfer knowledge, acquired from data-rich common classes, to semantically similar and yet data-hungry rare classes in order to obtain richer tail class representations. We provide an overview schematic of this idea in Figure~\ref{fig:teaser}. We leverage the fact that class prototypes and learned cosine classifiers provide two different representations of class cluster centres in feature spaces, and use an attention mechanism to transfer and recompose learned features from classifiers to prototypes. Importantly, information is only transferred from common to rare classes, addressing the limited reliability of classifiers trained in few shot settings and affording easy introduction of new, unseen classes in the classification process.

Our proposed approach is not tied to a specific model architecture or training strategy as it leverages a pre-trained model. As a result, we additionally provide a detailed analysis of the impact of cosine classifiers on pre-trained backbones, as well as the impact of sampling strategies on prototype quality.  
Based on this analysis, we propose training recommendations to optimise the feature and classifier learning stage for better knowledge transfer. 

Our \textbf{contributions} can be summarised as follows:
\begin{itemize}
    \item A novel knowledge transfer mechanism that composes semantically similar, learned, common-class representations to obtain richer and more discriminative tail-class classifiers. 
    Performance gains of up to $5\%$ on rare classes, while maintaining common class performance, without any additional training requirements.
    \item A flexible solution that affords continual adaptation to new, unseen classes; as well as the ability to construct classifiers tailored for specific class groups.
    \item An analysis of popular training strategies (sampling methods, classifier type) with recommendations towards optimisation of our transfer learning task.
\end{itemize}

\section{Related work}
\label{sec:related_work}

\subsection{Long-tail recognition.} 
\label{sec:related_work:longtail}
Long-tail recognition has received increasing attention in recent years, 
and popular benchmarks for the task now exist~\cite{OLTR}. Pre-existing works have mainly focused on the class imbalance problem.
Solutions proposed to address the issue can be separated in three categories: a) data resampling, b) rebalancing loss functions and post processing, c) ensembling and routing strategies. 

\noindent\textbf{Data resampling} is one of the most natural ideas when attempting to learn using imbalanced data. The idea involves oversampling less common classes~\cite{squareroot,chawla2002smote, decouple, iscen2021class} or undersampling the head classes by removing data~\cite{drummond2003c4}, which can lead to over or underfitting issues. Multiple oversampling strategies were studied in~\cite{decouple}, where it was shown that classifiers trained with oversampling achieve better overall performance, yet uniformly trained encoders exhibited higher robustness when applying different rebalancing techniques. 

More recently, resampling strategies have been mainly leveraged for classifier training. Two-stage training methods have been proposed~\cite{OLTR, decouple}, where a model encoder is first trained using uniform sampling, then a classifier is retrained with class balanced sampling. A two-branch siamese model was presented in~\cite{bbn}, with each branch receiving data samples following distinct distributions. The encoder and respective classifiers are trained using an adaptive combination of classification predictions from each branch. 
An adaptive sampling strategy was proposed in~\cite{balanced_softmax}, where the notion of an optimal common-to-rare class ratio is estimated via a meta-learning procedure, while a synthetic data generation process addressing class imbalance was recently proposed in~\cite{RSG}.

\noindent\textbf{Rebalancing loss functions.} Altering the training process through loss function definition constitutes a popular strategy, focusing on both representation learning and classification. Early works relied on reweighting samples in the cross-entropy loss~\cite{weightedCE}, achieving mixed results. The focal loss, which has shown great promise in object detection settings~\cite{focalloss}, also achieved limited improvement. Recent approaches involve a combination of rebalancing loss functions and post processing. 
A causality based strategy was introduced in~\cite{causalnorm}, relying on post training adjustment of normalised classifiers to reduce the impact of biased momentum-based training. 
Ren \etal~\cite{balanced_softmax} show that the softmax function yields biased gradient estimations, and introduce a balanced softmax alternative. 
Zhang \etal~\cite{disalign} and Hong \etal~\cite{lade} propose recalibration and postprocessing strategies to address the label distribution shift between training and test datasets. 
Zhong \etal~\cite{mislas} rely on label smoothing and mixup strategies, while Cui \etal~\cite{paco} combine cross-entropy with supervised contrastive learning, achieving stronger representations at the cost of more expensive training. 
Finally, closest to our work, Samuel \etal~\cite{samuel2021distributional} propose to rely on a prototype-based auxiliary loss, where class prototypes are computed at the beginning of every epoch during training. Introducing learnable class specific uncertainty parameters and weights yields stronger backbone representations, yet classification still relies on learned classifier weights. 

\noindent\textbf{Routing and ensembling} has been considered in multiple recent works~\cite{ride, tade, ace, reslt,iscen2021class}. The key assumption is that relying on multiple experts can afford the learning of more robust classifiers. Iscen \etal~\cite{iscen2021class} train a set of classification models, and then distill their knowledge to a new student model. Wang \etal~\cite{ride} propose to train multiple classifiers with a diversity loss, then learn a routing mechanism, where ambiguous predictions are redirected to a different expert. Cai \etal \cite{ace} propose to improve on this strategy by training different experts optimised for specific class groups. Further to this, Zhang \etal~\cite{tade} propose to learn ensembling weights in an unsupervised manner at test time via contrastive learning, in order to adapt to variable testing distributions. Ensembling strategies such as these tend to achieve superior performance, which can be partly attributed to increased capacity: each expert learns a different set of features in the last two blocks of ResNet models, consistently increasing model capacity as new experts are introduced, leading to 
scalability concerns. Finally, rather than learning higher capacity models, Cui \etal~\cite{reslt} propose to separate the parameter space into blocks dedicated to recognising different subsets of class groups, which are then aggregated at inference time. 

\paragraph{}
Very few approaches listed here introduce a mechanism that addresses the lack of training data at the tail and the risks of overfitting. Our work focuses on this issue, and is not tied to a rebalancing strategy. Therefore, it provides complementary benefits to approaches that focus on backbone training improvements\cite{paco,samuel2021distributional,balanced_softmax}, as discussed above.

\subsection{Few-shot learning}
\label{sec:related_work:fsl}

Also related to our work, few-shot learning focuses on developing methods specifically to address data scarcity related problems. These methods are typically developed with a sole focus on the few-shot classes, in artificial settings where all classes of interest have the exact same training data distribution, and models are pre-trained on a different set of base classes on which performance is rarely evaluated \cite{miniim, metadataset}. Popular FSL approaches typically rely on either distance based approaches \cite{protonets, dynamicFSL, imprintedweights}, or meta-gradient based techniques~\cite{MAML,lee2019meta,leo} which aim to quickly adapt models using only few gradient updates. 

Our work relates to distance based approaches, which commonly 
rely on the concept of class prototypes. Prototypes provide an alternative to classifier learning and are directly computed from the data using pre-trained encoders instead of being learned~\cite{protonets}. Recent works have proposed to combine prototypes with cosine classifiers \cite{imprintedweights, dynamicFSL} to facilitate training on base classes. Closer to our work, \cite{dynamicFSL} also sought to evaluate performance on base classes, and propagated classifier information during the rare classifier training process via an attention mechanism with rare training samples. Their approach however, still suffers from issues inherent with FSL research, as it relied on episodes to train and evaluate rare classifiers.

\section{Methodology}
\label{sec:ktc}

\subsection{Model pre-training}
\label{sec:ktc:pretrain}

Consider a training dataset $\mathcal{D} = \{\mathcal{X}, \mathcal{Y}\}$ comprising $N$ classes, training images $\mathcal{X}$ and corresponding class labels $\mathcal{Y}$. We assume this dataset is long-tail distributed, with $n_R$ rare classes defined as classes with the least amount of training data available (\eg less than $20$ samples). Further to this, we consider a classification model $\mathcal{M} = \{f_{\theta},W\}$ comprising a feature encoder backbone $f_{\theta}$, and classifier weights 
$W \in \mathds{R}^{F{\times}N}$, 
where $F$ is the feature dimension. 

Importantly, we define classifier $W$ as a cosine classifier, characterised by the fact that classification predictions are based on distance, rather than decision boundaries. More formally, for an input image $\mathbf{x}$ with feature representation $f_{\theta}(\mathbf{x})$, class predictions are computed as: 
\begin{align}
\label{eq:cosineclassif}
    q(y=i| \mathbf{x},\theta, W) = \frac{\exp\left\{S(f_{\theta}(\mathbf{x}),w_i) \right\}}
    {\sum_{k \in 1\dots N} \exp\left\{ S(f_{\theta}(\mathbf{x}),w_k )\right\}} \\
    \text{with } S(f_{\theta}(\mathbf{x}),w_i) = \frac{f_{\theta}(\mathbf{x})}{\norm{ f_{\theta}(\mathbf{x})}} \cdot \frac{w_i^{\top}}{\norm{w_i}} \cdot s
\end{align}
where $w_i$ is the $i^{th}$ column of classifier $W$, corresponding to class $i$'s center; and $s \in \mathbb{R}^{+}$ 
is a parameter commonly used at training time to obtain sharper softmax distributions and facilitate the training process.

Relying on models with the cosine similarity is crucial for our knowledge transfer process. The key advantage of training a distance based classifier is that it pushes semantically similar classes to be close to each other in feature space \cite{dynamicFSL}, facilitating inter-class knowledge transfer and feature comparison. Additionally, training models with fixed norm outputs has been observed to yield more stable training, and well clustered features \cite{wang2020understanding}. 

In subsequent sections, we assume a cosine classifier model $\mathcal{M}$ has been trained on dataset $\mathcal{D}$ using a standard cross-entropy loss. In our experiments, we will further evaluate and discuss the impact of using a cosine classifier, in addition to training settings to minimise the imbalance problem during this stage. We further note that while we consider a cross-entropy loss for simplicity and ease of evaluation, more complex rebalancing loss functions could be considered as well \cite{balanced_softmax,lade}.

\subsection{Compositional knowledge-transfer classifier}
\label{sec:ktc:cktc}

The cosine classifier provides learnable class representations that can be geometrically interpreted as class cluster centers in feature space, as they rely on cosine distance to determine classification assignments. These learned classifiers bear striking conceptual similarity to class prototypes, 
which are an alternative, learning-free strategy to obtain class cluster centres\cite{protonets}. 
They are typically computed directly from the training data as the average feature representation of all training samples in a given class: 
\begin{equation}
    p_i =  \frac{1}{n_i}\sum_{x_j \in \{\mathcal{X} \vert y_j = i  \}} \frac{f_{\theta}(x_j)}{\norm{ f_{\theta}(x_j)}}
\end{equation}
where $n_i$ is the number of training samples available for class $i$, and $y_j$ is the class index of sample $x_j$. Computed prototypes can then be used identically to learned cosine weights by replacing $w_i$ with $p_i$ in Equation \ref{eq:cosineclassif}.

\begin{figure}[t]
    \centering
    \includegraphics[width=\linewidth]{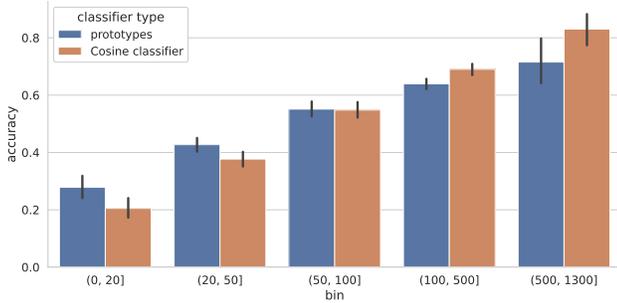}
    \caption{Comparison of class prototype and learned classifier weights accuracies with respect to the amount of class specific training data available. Validation accuracy from a model trained on the ImageNet-LT dataset.}
    \label{fig:protovsclassif}
\end{figure}

Typically, class prototypes have higher performance on rare classes, while learned cosine classifiers achieve the strongest common class accuracy (see Figure \ref{fig:protovsclassif}). One key takeaway is that both provide two class representations with the same conceptual interpretation (feature cluster centres), favouring different class groups.

Based on this observation, we propose to leverage both prototypes and cosine classifiers to construct a higher quality classifier. Our objective is to transfer knowledge acquired via learning from data-rich common classes towards semantically similar rare classes. We take inspiration from the Neural Turing Machines' \cite{NTM} reading operation, which reads and combines memory vectors according to cosine-similarity based coefficients. Using class prototypes as representation anchors due to their higher few-shot performance, we seek to identify common characteristics across classes that we recombine towards achieving high quality representation on rare classes. Using a prototype-to-classifier attention mechanism, we compute, for a given class $i$, a compositional class representation 
$w_i^{\text{kt}}$ 
that recombines relevant classifier knowledge from semantically similar classes:
\begin{align}
w^{\text{kt}}_i(k) &= 
    \sum_{j} \alpha_{ij} \frac{w_j}{\norm{w_j}} \\
    \mathbf{\alpha}_{ij} &= \text{softmax} \left (\frac{p_i}{\norm{p_i}} \cdot \frac{w_j^{\top}}{\norm{w_j}} \cdot\tau\right)  \\
    w_j \in W(k) &= \{w_l \vert l \in 1, \dots, N \text{, if } n_l > k   \}
\end{align}
where $\tau$ is a temperature parameter introduced to increase distribution sharpness and $k$ defines the minimum number of training samples available per class below which classifier knowledge is not transferred. 
This second parameter is crucial towards improving few-shot representations, allowing us to effectively transfer information from common to rare classes without propagating potentially unreliable few-shot features.
We compute our final classifier as:
\begin{equation}
     w^{\text{h}}_i(k) = 
     \frac{w^{\text{kt}}_i(k)}{\norm{w^{\text{kt}}_i (k)}}  
     + \frac{p_i}{\norm{p_i}} 
     + \frac{w_i}{\norm{w_i}} \label{eq:attention} 
\end{equation} 

\begin{figure}[t]
    \centering
    \includegraphics[width=1\linewidth]{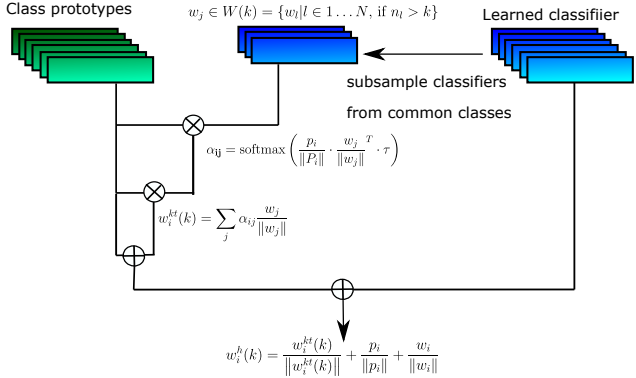}
    \caption{Illustration of the knowledge transfer classifier construction. See text for details. }
    \label{fig:ktfig}
\end{figure}

As class representations can be viewed as class centroids, 
$w^{\text{h}}_i(k)$ 
can be interpreted as the barycentric position between our multiple estimated centres (learned classifier, class prototypes and knowledge transfer component). The process is illustrated in Figure \ref{fig:ktfig}. We highlight one more time that this stage does not involve any additional training procedure, with classifiers and model weights obtained from model $\mathcal{M}$ after standard training. 

\paragraph{Ensembling strategy.}
Transferring knowledge from common to rare classes can significantly improve the discriminative power of few-shot classifiers. 
However, it can also reduce the positive biases inherent to trained classifier and reduce accuracy on common classes. Our knowledge transfer strategy affords flexibility on which class group classifiers are optimised for. 
Indeed, setting $k$ in Equation \ref{eq:attention} at a low value will yield classifiers dedicated to enhancing few-shot classes. Increasing this value will shift the focus to medium shot classes, as knowledge is transferred only from the most common classes. 
We can therefore create multiple prototype/cosine hybrid classifiers designed to achieve maximum performance on a specific class group. In light of this observation, we propose an ensembling strategy that combines predictions achieved with different classifiers optimised for different class groups: 
\begin{equation}
    q(y=i| \mathbf{x}) = \frac{1}{N_k}\sum_k q(y=i| \mathbf{x},\theta, W^h(k))
\end{equation}
Where $N_k$ is the number of ensembled classifiers.
A straightforward option is to choose the values of $k$ so as to build one classifier per class group: for example many-, medium- and few-shot. 

We highlight that we train a \emph{single classifier}, and consequently generate a set of new classifiers $W^h(k)$ (varying $k$ values) from the \emph{same set} of learnt, fixed classifier weights and prototypes. We therefore gain ensemble-esque benefits, via diverse classifiers, at almost no additional cost.

\paragraph{Continual adaptation.}
Last but not least, a noticeable advantage of our strategy, which does not rely on few-shot classification weights for knowledge transfer, is the ability to extend classifiers to new, previously unseen classes seamlessly. Indeed, this can be achieved with no additional training by computing new class prototypes, and has the potential to provide a strong, practical baseline. To achieve this, we propose a continual classifier estimation that does not rely on availability of few-shot classifiers as: 
\begin{equation}
\label{eq:continual}
    w^{\text{hc}}_i(k) = 
    \frac{w^{\text{kt}}_i(k)}{\norm{w^{\text{kt}}_i(k)}}  
    + \frac{p_i}{\norm{p_i}}
\end{equation}
We will show empirically that this approach can achieve competitive performance, and allows to integrate previously unseen classes seamlessly.

\section{Results}
\label{sec:exp}

\subsection{Experimental setup}
\label{sec:exp:setup}

We evaluate our approach on two standard long-tail recognition benchmarks: \textbf{ImageNet-LT} and \textbf{Places-LT}  \cite{OLTR}. In this section, we provide information on the dataset and our implementation details. Common to both datasets, we set the temperature parameter for our knowledge transfer step to $\tau=10$.
Unless specified otherwise, our models are trained with square-root sampling, and with a cosine classifier with learnable scale parameter initialised at 16 (parameter value as used in \cite{OLTR, causalnorm}). We provide an analysis of these two training components in our experiments. 
For ImageNet, we train a ResNext50 following \cite{decouple}.
For the Places dataset, we follow standard practice and use weights pre-trained on the full ImageNet dataset to initialise our model. We consider both supervised pre-training (standard) and unsupervised weights obtained using the SimCLR method \cite{simclr}, as well as ResNet101 and ResNet152 architectures. 
Detailed description of datasets and training parameters are provided in the supplementary material. 

Following standard evaluation protocols, we report classification accuracy over the entire test datasets, as well as over class groups defined as ``Many-shot'' ($n \geq 100$), ``Medium-shot'' ($ 20 < n < 100$) and ``Few-shot'' ($n \leq 20$).

\begin{table*}[t]
\centering
\caption{Classification accuracies on ImageNet-LT. All methods use a ResNext50 backbone. * models trained with a normalised classifier.}
\label{tab:INetresults}
\resizebox{0.75\linewidth}{!}{\begin{tabular}{l|l|c|c|c|c}
\toprule
Method & Classifier   type & Many-shot & Medium   shot & Few   shot & Total \\
\hline
\multirow{5}{*}{Ours*} & Cosine classifier & \bf68.5 & 45.6 & 20.6 & 50.9 \\
 & Prototypes & 63.5 & 49.1 & 27.4 & 51.6 \\
  & Prototypes + classifier & 67.0 & 48.7 & 25.2 & 52.4 \\
& $w^{\text{h}}(20)$ & 64 &  44.4 & \bf 48.2 &  52.7 \\
 &  $w^{\text{h}}(100)$  & 50.6 & \bf 57.8  & 36.8 &  51.9 \\
  &  ensemble$\left( w^{\text{h}}(0), w^{\text{h}}(20), w^{\text{h}}(100) \right)$&  63.2 &  52.1 &  36.9 & \bf 54.2 \\
 \hline
LWS~\cite{decouple}&  & 60.2 & 47.2 & 30.3 & 49.9 \\
BALMS~\cite{balanced_softmax} &  & 62.2 & 48.8 & 29.8 & 51.4 \\
Causal norm*~\cite{causalnorm}  &  & 62.7 & 48.8 & 31.6 & 51.8 \\
RSG*~\cite{RSG}  & & 63.2 & 48.2 & 32.3 & 51.8 \\
LADE~\cite{lade} &  & 62.3 & 49.3 & 31.2 & 51.9 \\
Disalign*~\cite{disalign} & & 62.7& 52.1 &31.4& 53.4  \\

\bottomrule
\end{tabular}}
\end{table*}

\subsection{Comparison to state of the art methods}
\label{sec:exp:sota}

It has been shown that backbone capacity and additional training losses can substantially improve backbone quality and consequently classification performance \cite{paco,decouple,samuel2021distributional}. As our approach is independent from the backbone training process, we train our model using the setting that matches most pre-existing approaches. We use a standard cross entropy setting and evaluate our method with respect to state of the art approaches using the same backbone architecture and training regime for fair comparison. Notably, we highlight methods leveraging normalised classifiers, as these tend to yield superior backbones (see Section \ref{sec:exp:backbone}). 

We provide detailed results on the ImageNet dataset in Table \ref{tab:INetresults}, comparing individual knowledge transfer classifiers ($k=20$, $k=100$) with an ensemble strategy ($k=0$, $k=20$, $k=100$). We can see that the ensembling strategy achieves the most robust performance, increasing or matching state of the art accuracy in all class groups. We achieve the most substantial performance gain on the few-shot classes, improving accuracy by at least $5\%$ to $16\%$ with respect to competing methods. 
As individual classifiers are tailored for a specific class group, combining them via ensembling strategy allows to maintain robust performance on all class groups while significantly improving the performance on rare classes.

We additionally provide results for the ensemble strategy ($k=0$, $k=20$, $k=100$) the Places dataset in Table \ref{tab:placesresults} (more detailed results are provided in the supplementary material). We provide results on ResNet101 with different pre-trained weights, as well as ResNet152 with supervised initialisation\footnote{Deeper architectures such as ResNet152 are not commonly used in~self-supervised learning settings.} for fair comparison with state of the art methods. 
Our first observation is that we achieve stronger results on ResNet101 despite the reduced capacity. We hypothesize that stronger overfitting issues (notably due to the square-root sampling process) can be encountered with the larger capacity backbone, and that unsupervised pre-training achieves stronger performance due to its higher compatibility with our cosine classifier. 
Nonetheless, we note that we achieve overall accuracy on par with state of the art models (ResNet152), and outperform all methods using our ResNet101 initialised with an unsupervised backbone. Similarly to the ImageNet dataset, we obtain the largest performance gains on few-shot classes, outperforming state of the art methods on few-shot classes by at least 2.8\% in all settings.

\begin{table*}[t]
\centering
\caption{Classification accuracies on the ImageNet-LT dataset in the continual setting. }
\label{tab:continual}
\resizebox{0.9\linewidth}{!}{\begin{tabular}{l|l|c|c|c|c}
\toprule
Method & Classifier   type & Many-shot & Medium   shot & Few   shot & Total \\
\hline
\multirow{3}{*}{Full training} & Cosine classifier & \bf68.5 & 45.6 & 20.6 & 50.9 \\
 & Prototypes & 63.5 & 49.1 & 27.4 & 51.6 \\
&  $w^{\text{hc}}(20)$ & 61.7 &  44.7 & \bf 48.3 &  51.8 \\
 &  $w^{\text{hc}}(100)$ & 51.1 & \bf 55.3  & 37.2 &  51.0 \\
  
   & ensemble$\left( prototypes,w^{\text{hc}}(20), w^{\text{hc}}(100) \right)$  &  61.4 &  51.1 &  38.3 & \bf53.2 \\ 

\hline
 \multirow{3}{*}{Few-shot excluded}  & Cosine classifier & \bf66.0 & 40.8 & N/A & N/A \\ 
 & Prototypes & 61.8 & 43.5 & 26.2 & 48.1 \\
&  $w^{\text{hc}}(20)$ & 57.3 &  37.7 & \bf 39.3 &  45.6 \\
 &  $w^{\text{hc}}(100)$ & 50 & \bf 50.2  & 25.8 &  46.6 \\
  
   & ensemble$\left( prototypes,w^{\text{hc}}(20), w^{\text{hc}}(100) \right)$  &  58.9 &  45.0 &  32.2 & \bf48.6 \\

\bottomrule
\end{tabular}}
\end{table*}

\begin{table}
\centering
\caption{Classification accuracies on the Places-LT dataset. *~refers to models trained with a normalised classifier.}
\label{tab:placesresults}
\resizebox{\linewidth}{!}{\begin{tabular}{l|c|c|c|c}
\toprule
Method & Many-shot & Medium   shot & Few   shot & Total \\
\midrule
\multicolumn{5}{c}{ResNet101 backbone} \\
\midrule
Ours* - supervised init & 40.8 & 40.1 & 34.9 & 39.3 \\
\hline
Ours* - unsupervised init & \bf41.6 & 41.4 & 35.1 & \bf40.2 \\
\hline
Disalign R101~\cite{disalign}  & 39.1& \bf42.0 & 29.1&  38.5  \\
\midrule
\multicolumn{5}{c}{ResNet152 backbone} \\
\midrule
Ours* - supervised init &  39.7 & 41.0 & \bf34.9 & \bf39.2 \\

\hline
LWS \cite{decouple}&   40.6 & 39.1 & 28.6 & 37.6 \\
BALMS~\cite{balanced_softmax} &   42.0 & 39.3 & 30.5 & 38.6 \\
LADE \cite{lade}&   \bf42.8 & 39 & 31.2 & 38.8 \\
RSG*~\cite{RSG}  &  41.9 & 41.4 & 32.0 & \bf39.3 \\
Disalign R152~\cite{disalign} &  40.4& \bf42.4 & 30.1&  \bf39.3  \\
\bottomrule
\end{tabular}
}
\end{table}

\subsection{Continual adaptation}
\label{sec:exp:adapt}

Next, we aim to evaluate our model's ability to achieve good recognition performance in settings where some classes have never been seen before. To this end, we train a model on the ImageNet-LT dataset, removing all few-shot classes ($n\leq 20$, 146 classes) from the training set. We refer to this model as our continual model in future references.
We evaluate the continual classifier $w_i^{hc}$ as introduced in Equation \ref{eq:continual} on both a model trained on the full dataset, and our continual model. We report results in Table \ref{tab:continual}, considering classifier accuracy, prototype accuracy. $w_i^{hc}$ for $k=\{20,100\}$ and an ensemble of both classifiers with class prototypes. We can see that our continual model still achieves a robust performance despite the fact that the backbone wasn't trained to separate common from rare classes. We note that we still achieve very good recognition performance on rare classes, outperforming almost all state of the art methods on this class group (see Table \ref{tab:INetresults}) even though they were never seen during model training. 

\subsection{Backbone training}
\label{sec:exp:backbone}

In this section, we focus on the standard learning setting using a cross-entropy loss, and aim to evaluate the impact of different training settings on backbone quality.  More specifically, we evaluate the influence of data sampling strategies and type of classifier. 
Our main focus is to evaluate the quality of trained backbones with respect to prototype estimation and few-shot class representation. 

In order to identify the preferred training strategy, we evaluate models trained using \textbf{a)} trained classifiers, \textbf{b)} prototypes computed using the training set (training prototypes), \textbf{c)} prototypes computed using the validation set (validation prototypes). We measure test accuracy for classification over all classes, as well as individual class groups only, aiming to evaluate backbones with regards to class representation quality, generalisation ability, and common class bias.
Lastly, in order to evaluate whether oversampling led to important memorising,  
we measure accuracy on the training set, using validation prototypes as classifiers. 
Models are trained on the ImageNet-LT dataset using the same parameters and architectures as described in our experimental setup section unless specified otherwise.

\noindent\textbf{Type of classifier.} 
Relying on a cosine classifier is a key condition in our approach to achieve successful knowledge transfer between classifiers and prototypes. Here, we compare the performance of models trained using cosine classifiers with the popular linear dot product classifier.  

\begin{figure}[t]
    \centering
    \includegraphics[width=\linewidth]{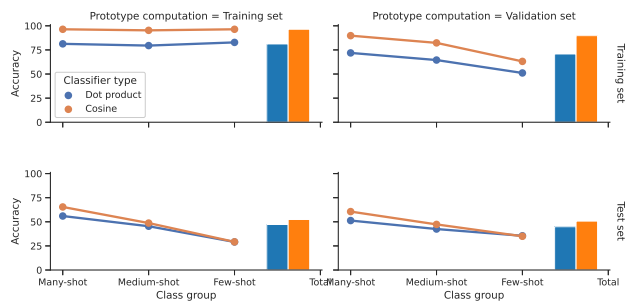}
    \caption{Influence of classifier type on class separability. Prototype-based prediction accuracy is computed with respect to the three different class groups on the training and test sets, with prototypes computed on training and validation sets.}
    \label{fig:classiftype}
\end{figure}

We provide classifier-based accuracies in Table \ref{tab:classif}, and prototype-based accuracies in Figure \ref{fig:classiftype}. We can see that usage of the cosine classifier significantly increases classification accuracy for all class groups. Prototype based accuracies show that we consistently achieve better class separability, with the exception of few-shot classes where performance is equivalent. This suggests that cosine classifiers facilitate model training, achieving better accuracy and separability faster. The use of a distance based classifier also appears to strongly benefit few-shot representations. Normalisation based classifiers, including cosine classifiers, were considered to be beneficial to reduce momentum induced bias in \cite{causalnorm}, and were also observed to yield consistently better performance in \cite{disalign}.

\begin{table}%
  \centering
    \caption{Classification accuracy using learned classifiers to evaluate the influence of a) classifier type, and b) sampling strategies.}
  \subfloat[\label{tab:classif} Classifier type]{
  \resizebox{\columnwidth}{!}{\begin{tabular}{l|c|c|c|c}
\toprule
 Classifier &Many-shot & Medium-shot & Few-shot & total \\
\hline
Dot product & 65.9 & 37.5 & 7.5 &  44.4\\
 Cosine & \bf69.2 & \bf43.0 & \bf15.4 & \bf49.2 \\
 \bottomrule
\end{tabular}}
  }%
  \qquad
  \subfloat[\label{tab:classifsamp} Sampling method]{
 \resizebox{\columnwidth}{!}{ \begin{tabular}{l|c|c|c|c}
\toprule
Sampling  &Many-shot & Medium-shot & Few-shot & total \\
\hline
Uniform & \bf69.2 & 43.0 & 15.4 & 49.2 \\
  Square   root & 68.5 & \bf45.6 & \bf20.6 & \bf50.9\\
  Class   aware & 64.1 & 38 & 14.2 & 44.7 \\
 \bottomrule
\end{tabular}}
  }

  \label{tab:backclassifs}%
\end{table}

\noindent\textbf{Sampling strategy.}
A simple way of rebalancing long-tail training datasets is to prevent the dominance of common classes during the training process. This can be achieved via sampling strategies when constructing training batches, with the aim to sample rare classes more often. Finding the right balance is not straightforward however, as oversampling rare classes can lead to overfitting and memorising. 
Different batch sampling strategies were proposed in \cite{decouple}, and evaluated with respect to different long-tail approaches. 
It was observed that oversampling rare classes lead to stronger classifiers after standard training. However, this performance gain was lost after applying classifier rebalancing strategies, with sample uniform sampling methods achieving the best performance. 

Here, we aim to further evaluate backbone encoders trained using different sampling strategies, and determine the optimal sampling strategy for our task. In addition, we aim to verify that previous observations still hold for cosine classifiers. We seek a combination of high quality encoder backbone and class prototypes. We evaluate backbones trained using cosine classifiers with identical parameters and three sampling strategies: a) \textbf{Sample uniform}: standard training setting, where images are uniformly sampled from the training database; b) \textbf{Class uniform} \cite{chawla2002smote}: uniformly sampling categories, then selecting a fixed number of samples per class (common practice is $4$ samples). This results in substantially oversampling rare classes; and c) \textbf{Square root}\cite{squareroot}: the probability of sampling images from class $j$ is $q_j = \frac{n_j^{1/2}}{\sum_{i=1}^N n_i^{1/2} }$. This increases the likelihood of sampling from rarer classes but does not dramatically alter the data distribution.

\begin{figure}[t]
    \centering
    \includegraphics[width=\linewidth]{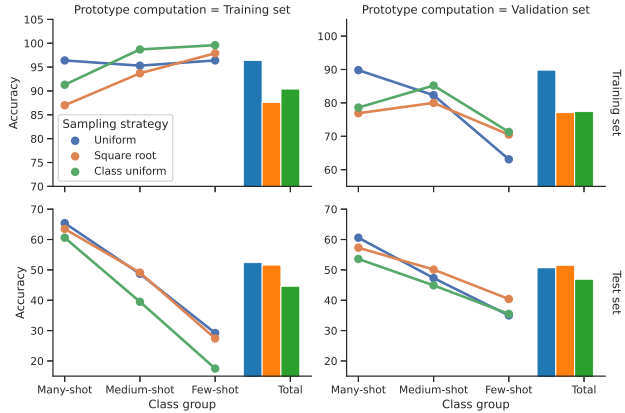}
    \caption{Influence of data sampling strategies on class separability. Prototype-based prediction accuracy is computed with respect to the three different class groups on the training and test sets, with prototypes computed on training and validation sets.}
    \label{fig:prototype_sampling}
\end{figure}

Classifier accuracy is reported in Table \ref{tab:classifsamp}, prototype classification accuracies are provided in Figure \ref{fig:prototype_sampling}, and class group level accuracy in Table \ref{tab:prototype_sampling_clgroup}. Our first observation refers to the class uniform sampling approach, which achieves the poorest performance, both with regards to classifiers and prototypes. Our experiments in Figure  \ref{fig:prototype_sampling} (top left) suggest overfitting, with performance on the training set being close to 100\% for both few and medium shot classes. The poorer performance compared to dot product classifiers \cite{decouple} could be linked to cosine classifiers enabling faster training. Second, square root sampling achieves the highest classification accuracy, with performance slightly reduced for many-shot classes. While square-root and uniform sampling have comparable performance on training prototypes, our experiments on validation prototypes (right column in Figure \ref{fig:prototype_sampling}) suggest that square root models provide better separability for medium and few-shot classes overall, which is crucial for our knowledge transfer process. This is additionally supported by Table \ref{tab:prototype_sampling_clgroup}, which shows, when considering classification within few-shot classes only, that square-root sampling achieves stronger accuracy. 
 
These observations highlight that sampling strategies can have a substantial impact on model performance, and that square-root sampling can provide better representations for our objective. We note however that the optimal sampling method could differ depending on data distribution, and recommend square-root sampling for datasets exhibiting a similar distribution to the ImageNet long-tail dataset.

\begin{table}[t]
\caption{Influence of data sampling strategies on class group prototype accuracy. Classification within each class group only.  }
\centering
\label{tab:prototype_sampling_clgroup}
\resizebox{\columnwidth}{!}{\begin{tabular}{l|l|c|c}

\toprule
Prototype source & Sampling  & Few-shot & Many-shot \\
\hline
\multirow{3}{*}{Training   set} & Uniform & 60.8 & \bf70.6 \\
 & Square   root & \bf61.9 & 69.9 \\
 & Class   aware & 52.1 & 65 \\
 \hline
\multirow{3}{*}{Validation   set} & Uniform    & 63.3 & \bf69.7\\
 & Square   root  & \bf67.9 &  68.1 \\
 & Class   aware  & 62 & 63 \\
 \bottomrule
\end{tabular}}
\end{table}

\section{Conclusion}
\label{sec:conclusion}

In this work, we introduce a novel approach for long-tail recognition focused on the tail's few-shot problem. We introduce a knowledge transfer strategy that recomposes learned classification features from data-rich common classes to improve rare class representations in a training-free manner. Our experiments on two standard benchmarks show that we achieve significant performance gains on rare classes, while maintaining strong performance on common classes. We further demonstrate our model's ability to integrate new, previously unseen classes with strong recognition accuracy, and provide an analysis of our model training strategy so as to facilitate knowledge transfer. 
Nonetheless, our approach is not tied to a specific backbone training approach, and further benefits could be obtained by combining our strategy with backbone training methods comprising auxiliary losses that optimise for class separability \cite{paco,samuel2021distributional}. 

\noindent\textbf{Limitations.}
One limitation of the knowledge transfer method, described in Sec.~\ref{sec:ktc}, is that performance on common classes may suffer. We mitigate this effect via ensembling, but additional options may be considered, such as dual classifiers or separate memory vectors. Prototypes, as class average representations, can lack the precision that learned classifiers provide. We therefore plan in the future to explore richer prototype representations \eg distribution-based or multiple prototypes per class \cite{IMP}. Currently, our idea relies on a training-free strategy, and cannot fully leverage new data for backbones in \eg continual scenarios. Training-based extensions can also be considered in future work.
Finally, our approach relies on the assumption that there exists classes with similar visual properties. Our knowledge transfer approach might not be successful in settings where this is not the case. As visual similarities are currently estimated via feature comparisons, introducing additional structured, prior knowledge (\eg knowledge bases) may further inform the crucial decisions of both when and if knowledge transfer should occur.

{\small
\bibliographystyle{ieee_fullname}
\bibliography{egbib}
}

\clearpage
\appendix 

\section{Datasets and implementation details}

\paragraph{ImageNet-LT}  \cite{OLTR} is a subset of the 
large-scale ImageNet dataset \cite{Imagenet}, subsampled such that class distributions follow a Pareto distribution with power value $\alpha=6$. The dataset contains 116k training images from 1,000 categories, with class cardinality
ranging from 5 to 1,280. The dataset is publicly available, and its usage is limited for research only (non-commercial or education purposes).

We trained our backbone encoder with parameters following the training setting most commonly used in the literature \cite{decouple}: we train ResNext50 models, with cosine classifiers, for 90 epochs, with weight decay $0.0005$, batch size $512$, and learning rate initialised at $0.2$ with cosine decay to 0. All sampling methods use identical parameter sets.  

\paragraph{Places-LT} is a subset of the large-scale scene classification dataset; Places~\cite{places} that is constructed in a similar fashion to the ImageNet-LT dataset~\cite{OLTR}. The dataset comprises of 62.7K training images from 365 categories with class cardinality ranging from 5 to 4980. The dataset is publicly available, and its usage is limited to research only (non-commercial or education purposes).

We trained a ResNet152, using supervised pre-trained weights, towards direct comparison with state of the art methods. Due to the lack of publicly available pre-trained ResNet152 models, we carry out our backbone analysis and additional experiments using an unsupervised initialisation of ResNet101 architectures. All pre-trained weights are obtained by leveraging the full ImageNet dataset according to standard practice. 

All models are trained following standard practice with regards to parameters commonly found in the literature~\cite{OLTR,lade}. We use a batch size of $128$, weight decay of $0.0005$. The learning rate is set to $0.001$ for the pre-trained backbone encoder, and $0.1$ for the cosine classifier with a cosine decay to $0$. We train ResNet152 models for 10 epochs, and ResNet101 models for 15 epochs.

\paragraph{}
Our method is implemented using PyTorch~\cite{pytorch}.

\section{Societal impact}
The potential benefits of low data regime tools often relate to reduction of data costs; collection, curation, storage and processing. Our approach in particular, can contribute to reducing recognition bias with regards to under-represented classes, that involve rare or otherwise difficult to acquire training samples. Furthermore, our approach allows to adapt pre-trained models to reduce biases or introduce new classes without additional training steps, consequently improving environmental impact. In terms of risks; making models both readily available and quickly accessible 
for novel tasks, at low data and training costs, to individuals without  
domain 
expertise, in combination with potentially increased susceptibility to subtle prediction failures, may increase the risk of both models and their outputs being used incorrectly.

\section{Additional results on the ImageNet dataset}

We provide an ablation experiment showing the impact of changing prototype and classifier roles in Eqs.~{4--6}.
Results are summarised in Table \ref{tab:attentionabl}, where the first row report our original results (prototype to classifier weights attention). 
We can see that the best performance is achieved in our chosen configuration. The decreased performance in other configurations is to be expected, as less reliable few-shot classifiers and/or many-shot prototypes are relied on more heavily in other configurations.

We provide additional results on the ImageNet dataset in Table~\ref{tab:INetresults}: we report performance using a model trained using uniform sampling, as well as a model trained using a balanced softmax loss~\cite{balanced_softmax} (\vs regular cross entropy loss). Our balanced softmax model is retrained using a cosine classifier, which explains our higher accuracy with regards to numbers reported in~\cite{balanced_softmax}.
It may be observed that our approach continues to improve performance on rare classes, and in particular we note the high performance on this class group using a balanced softmax classifier. We further note how backbones influence our final overall performance (\eg uniform \vs square root), highlighting the importance of training a high quality backbone. We state again that solutions that aim to learn better representations are complementary of our method, which focuses on handling the few-shot problem.

\begin{table}[t] 
\centering
\caption{Sensitivity results evaluating the impact of the prototype to classifier attention mechanism. $p\xrightarrow{}w$ corresponds to the setting described in Eq.~{4--6} of the manuscript, $w\xrightarrow{}p$ inverts classifier and prototype roles, $w\xrightarrow{}w$ and $p\xrightarrow{}p$ carry out self attention with only one kind of class representation.}
\label{tab:attentionabl}
\begin{tabular}{l|llll}
\toprule
 & Many-shot & Medium-shot & Few-shot & Total \\
 \hline
$p\xrightarrow{}w$ & 63.2 & \textbf{52.1} & \textbf{36.9} & \textbf{54.2} \\
$w\xrightarrow{}p$ & \textbf{65.0} & 50.1 & 28.9 & 52.9 \\
$w\xrightarrow{}w$ & 64.5 & 51.1 & 31.3 & 53.4 \\
$p\xrightarrow{}p$ & 62.1 & 50.1 & 31.4 & 52.1\\
\bottomrule
\end{tabular}
\end{table}

\begin{table*}[t] 
\centering
\caption{Classification accuracies on ImageNet-LT. All methods use a ResNext50 backbone. * models trained with a normalised classifier.}
\label{tab:INetresults}
\resizebox{0.85\linewidth}{!}{\begin{tabular}{l|l|c|c|c|c}
\toprule
Method & Classifier   type & Many-shot & Medium   shot & Few   shot & Total \\
\hline
\multirow{2}{*}{Uniform sampling} & Cosine classifier & \bf69.2 & 43.0 & 15.4 & 49.2 \\
  &  ensemble$\left( w^{\text{h}}(0), w^{\text{h}}(20), w^{\text{h}}(100) \right)$&  62.9 &  48.9 &  33.7 & \bf 52.2 \\
  \hline
\multirow{2}{*}{Balanced softmax} & Cosine classifier & \bf64.2 & 49.2 & 32.8 & 52.7 \\
  &  ensemble$\left( w^{\text{h}}(0), w^{\text{h}}(20), w^{\text{h}}(100) \right)$&  61.4 &  \bf 49.9 & \bf 38.8 & \bf 52.8 \\

\bottomrule
\end{tabular}}
\end{table*}

\section{Attention mechanism class selection}

\begin{figure*}[t]
    \centering
    \includegraphics[width=\linewidth]{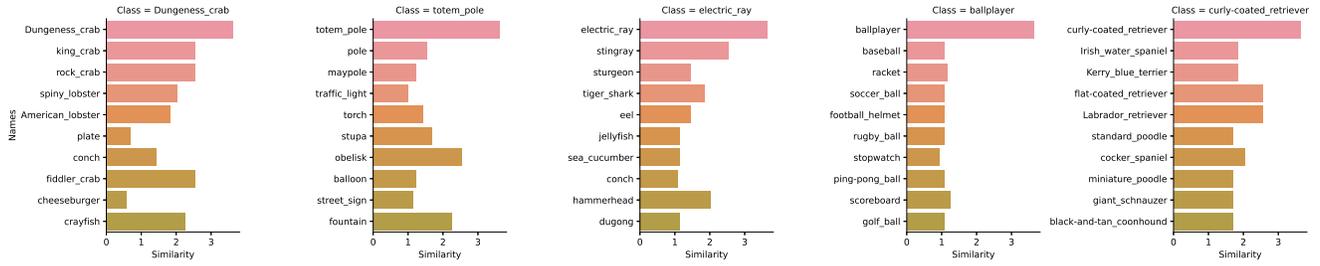}
    \caption{For five randomly selected classes, we report the ten nearest classes in terms of cosine similarity, with respect to their semantic similarity according to the WordNet taxonomy.}
    \label{fig:class_attn}
\end{figure*}

In this section, we evaluate our model ability to select semantically relevant classes via our attention mechanism. To this end, we show in Figure~\ref{fig:class_attn} the classes with top $10$ attention scores for five randomly selected classes. For simplicity, we consider $k=0$, such that all classes are treated identically. To evaluate semantic similarity, we plot class semantic similarity in the WordNet hierarchy by computing the Leacock-Chodorow Similarity~\cite{leacock1998combining}, which measures the shortest path distance between classes in the graph, while taking into account their depth in the taxonomy. We can see that classes with very similar categories are selected (i.e. dog breeds), allowing for transferring common properties across these classes.

\section{Visualizing the impact of knowlege transfer}
\begin{figure}

  \includegraphics[trim=3.1cm 3cm 2.5cm 2.6cm,clip,width=\linewidth]{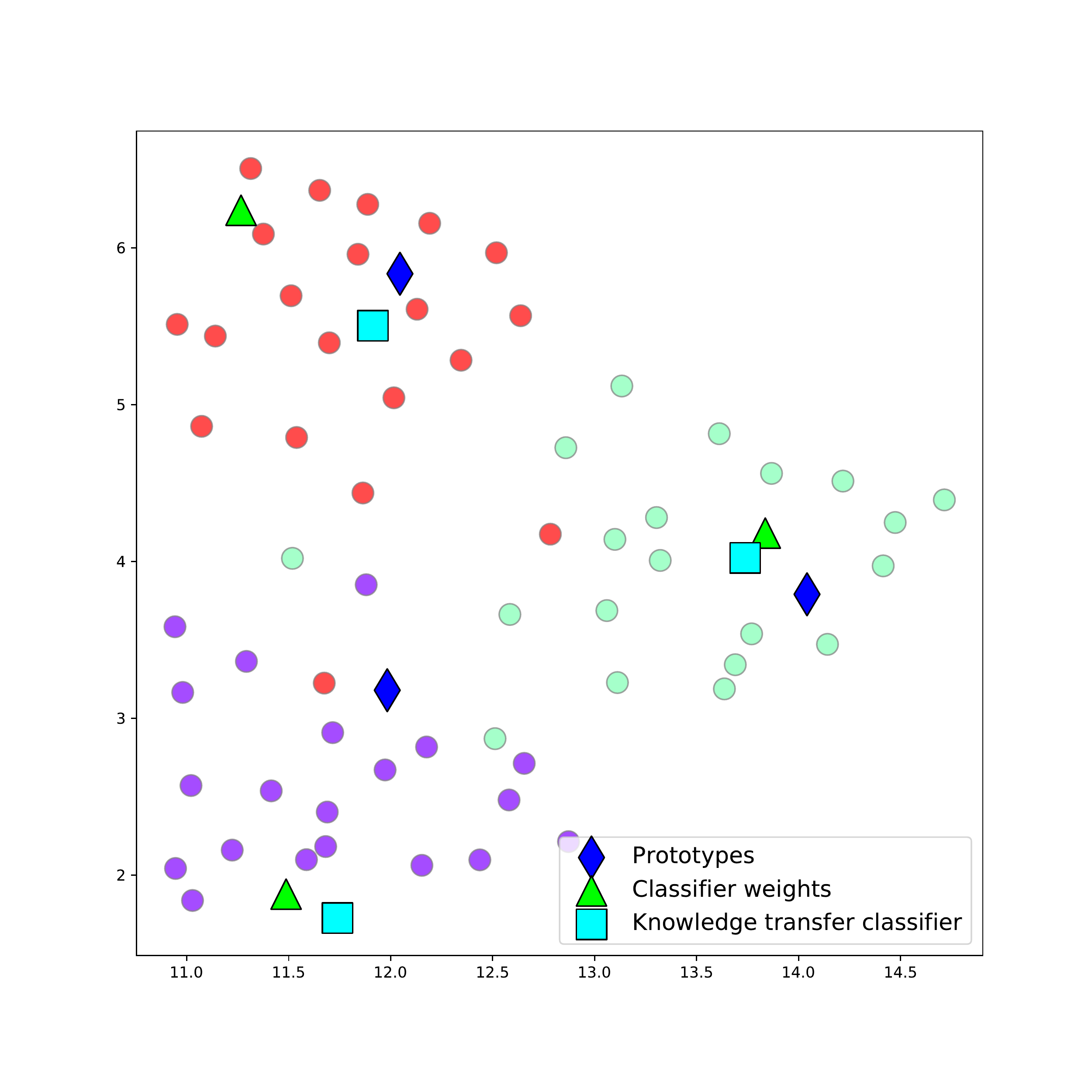}
  \caption{{UMAP representations of validation samples from a few-shot class [red cluster] and the top-2 closest classes in terms of knowledge transfer attention score [purple (many shot class) and green (medium shot) clusters]. For each class, we also plot prototypes, classifiers and final classifier.}\label{fig:TSNE}}
\end{figure}

Fig~\ref{fig:TSNE} shows the impact of knowledge transfer on class representations, visualising how our knowledge transfer process adjusts class representations. We can see that after transfer, the few-shot class representation is pushed towards more accurate class prototypes, while common class representations remain close to their learned classifier.
We also note that it illustrates how prototypes and trained classifier representations can differ, highlighting the advantage of combining these two representations. 

\section{Additional results on the Places dataset}

\begin{figure*}
    \centering
    \includegraphics[width=\linewidth]{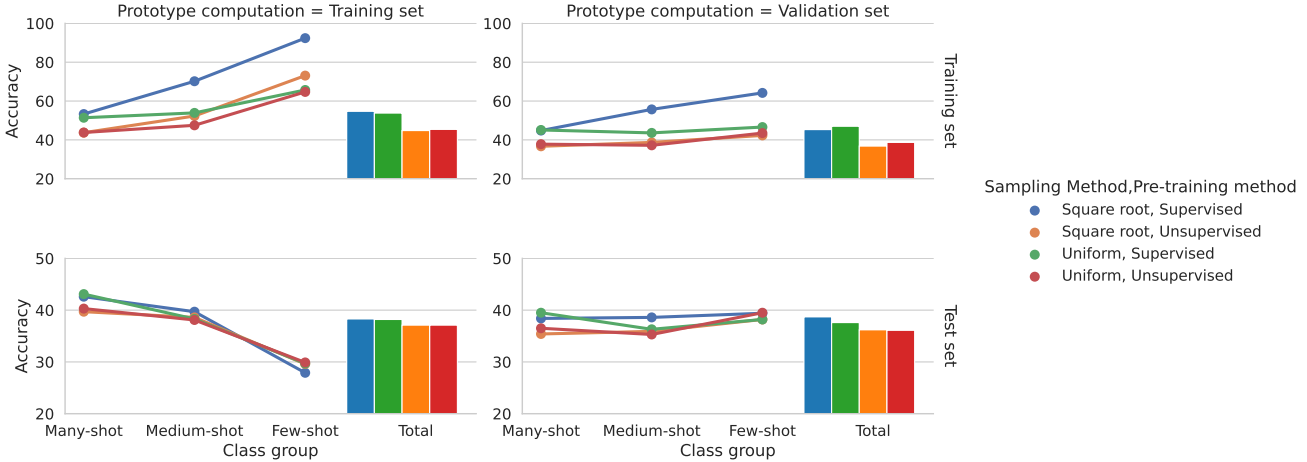}
    \caption{Influence of data sampling strategies and pre-training weight choice on class separability for the Places-LT dataset. Prototype-based prediction accuracy is computed with respect to three different class groups on the training and test sets, with prototypes computed on training and validation sets.}
    \label{fig:places_backbone}
\end{figure*} 

\subsection{Backbone analysis}

Further to our ImageNet-LT backbone analysis, we provide here further study on training strategies of interest, additionally for the places dataset. We carry out this study on the ResNet101 backbone, and consider: a) the sampling strategy (square root or uniform), and b) the choice of pre-trained weights (supervised or unsupervised pre-training on ImageNet). Due to the smaller size of the dataset, experiments in the literature on Places-LT typically initialise the model backbone encoder with weights pre-trained, in a supervised manner, on the entire ImageNet dataset.  

Here, we additionally consider usage of an unsupervised initialisation, with a model pre-trained on ImageNet using the SimCLR~\cite{simclr} contrastive learning method. Unsupervised pre-training has been shown to achieve superior transfer learning performance in certain circumstances~\cite{ericsson2021well}. In addition, our key incentive is the fact that, in contrast to standard supervised pre-training, SimCLR relies on a \emph{normalised, distance-based} representation learning process, which has higher compatibility with our cosine classifier strategy. 

To evaluate which training strategy yields higher quality backbones, we consider the same initialisation criteria discussed in the main manuscript: We compute training and validation prototypes, and measure accuracy on both the training (to evaluate underfitting and memorisation), and test sets. 
Our analysis is reported in Figure~\ref{fig:places_backbone}. Firstly, in terms of sampling strategies, we note a limited impact overall, with square root yielding higher quality prototypes with supervised initialisation, and uniform sampling having a very slight edge when using an unsupervised initialisation. 
We can see that models trained using a supervised initialisation achieve better training accuracy, suggesting underfitting when using an unsupervised initialisation. We note that the supervised models achieve better performance on classes with sufficient data, while better performance is obtained for few-shot classes when using unsupervised models.

Further to this, we seek to analyse the compatibility between classifiers and prototypes, as well as the sharpness of our attention mechanism. To evaluate this we compute, for each class prototype, the cosine distance to all cosine classifiers weights. Firstly, we evaluate the number of class prototypes which are not closest to their corresponding cosine weight (\ie the prototype from class $A$ is closer to the classifier from class $B$, instead of the classifier from class $A$). Selecting the wrong class suggests poorer compatibility, reducing the accuracy of our attention mechanism and knowledge transfer process. We report this result in Table~\ref{tab:counts}, showing that square root models achieve better compatibility, and that the supervised model yields the worst results.  

\begin{figure}[t]
    \centering
    \includegraphics[width=\linewidth]{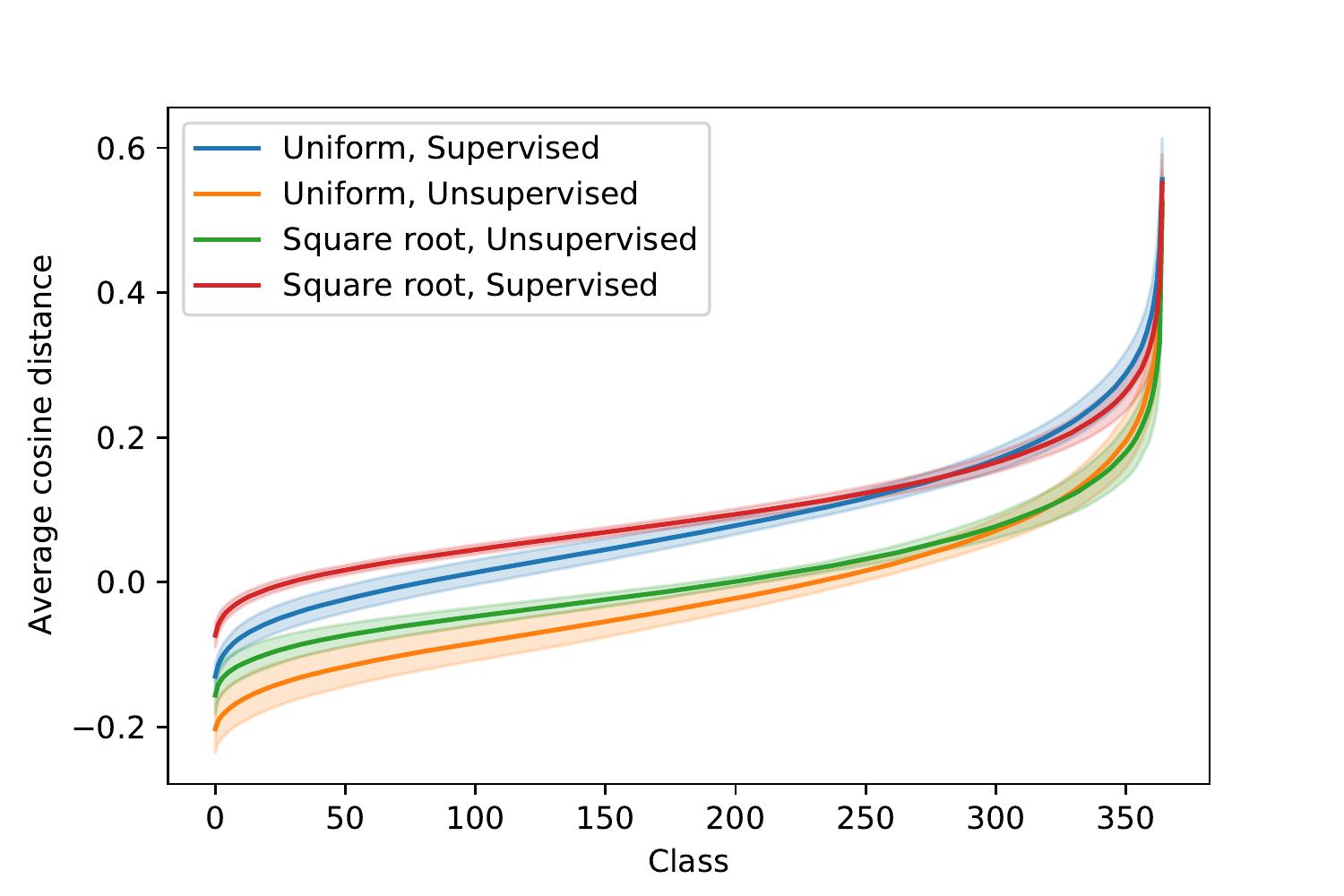}
    \caption{Influence of data sampling strategies and pre-training weight choice on prototype compatibility and our attention mechanism. We report, for sampling and pre-training weights considered, the average over all prototypes of the sorted cosine similarity between a class prototype and cosine classifier weights of all classes. }
    \label{fig:places_distances}
\end{figure}

In addition to this, we evaluate how sharp cosine similarity distributions are between a given prototype and classifier weights. Intuitively, one seeks sharp distributions, with only a handful of classes possessing high similarity with the prototype of interest so as to only transfer knowledge from the most relevant classes. This is visualised in Figure~\ref{fig:places_distances}, where it may be observed that backbones relying on unsupervised weights tend to obtain sharper, and therefore more selective distributions. While square root and uniform sampling appear to behave similarly, we note that uniform sampling yields slightly sharper distributions, giving it a slight edge again.  

In light of this, we expect models initialised with unsupervised weights to achieve stronger performance due to their higher prototype / classifier compatibility. 

\begin{table}%
  \centering
    \caption{Number of classes where prototypes are not closest to their corresponding cosine classifier for multiple backbones.} 
    \label{tab:counts}
 \begin{tabular}{l|c}
\toprule
Backbone & Mismatch count \\
\hline
Uniform, Supervised & 39 \\
Uniform, Unsupervised & 17 \\
Square root, Supervised & 0 \\
Square root, Unsupervised & 0 \\
 \bottomrule
\end{tabular}
\end{table}

\subsection{Ablation experiments} 

We provide additional detailed results on the Places-LT dataset for all studied ResNet101 backbones in Table~\ref{tab:placesresults}. As was conjectured in the previous section, we achieve better performance using an unsupervised initialisation, and, remarkably, that performance is equivalent between the two sampling strategies with $40.2$ total accuracy.

Interestingly, the backbone exhibiting the poorest performance combines a supervised initialisation with uniform sampling, resulting in the weakest performance in almost all settings. Nonetheless, we note that all backbone configurations outperform state of the art method~\cite{disalign} when employing a ResNet101 backbone.

\begin{table*}
\centering
\caption{Detailed classification accuracies and ablations on the Places-LT dataset. Bold numbers highlight the best performing classifier type per backbone.}
\label{tab:placesresults}
\resizebox{0.8\linewidth}{!}{\begin{tabular}{l|c|c|c|c}
\toprule
Method & Many-shot & Medium   shot & Few   shot & Total \\
\midrule
\multicolumn{5}{l}{\bf Supervised initialisation, square root} \\
\midrule
  Prototypes & 42.6 & 39.7&27.9  & 38.3 \\
  Cosine classifier & \bf47.5& 35& 19.7&36.3  \\
  Classifier + prototypes & 46.6 &  37.2 & 22.1 & 37.4 \\
   $w^h(20)$ & 41.6 &  30.4 & \bf 44.8 & 37.5 \\
   $w^h(100)$ & 27.8 & \bf46.9   & 33.2 & 37.1  \\
  ensemble($w^{hc}(20)$,$w^{hc}(100)$) (\bf continual) & 36.2  & 37.8  & 41.9  & 38.1 \\

ensemble($w^h(0)$,$w^h(20)$,$w^h(100)$) & 40.8 & 40.1 & 34.9 & \bf 39.3 \\
\midrule
\multicolumn{5}{l}{\bf Unsupervised initialisation, square root} \\
\midrule
  Prototypes & 39.7 & 38.6& 29.6 & 37.1 \\
  Cosine classifier & \bf 48.4& 33.8& 18.2& 35.8\\
  Classifier + prototypes & 45.7 & 38.4 & 25.1 & 38.2\\
   $w^h(20)$ & 42.3 & 32.6 & \bf44.3 & 38.6 \\
   $w^h(100)$ & 31.0 & \bf 47.2 & 34.2 & 38.6\\
 ensemble($w^{hc}(20)$,$w^{hc}(100)$) (\bf continual) & 38.5  & 38.4  & 39.5  & 38.6 \\
ensemble($w^h(0)$,$w^h(20)$,$w^h(100)$)  & 41.6 & 41.4 & 35.1 & \bf40.2 \\
\midrule
\multicolumn{5}{l}{\bf Supervised initialisation, uniform} \\
\midrule
  Prototypes & 43.1 & 38.3 & 29.7 & 38.2 \\
  Cosine classifier & \bf 48.1 & 25.7 & 10.0 & 30.5 \\
   Classifier + prototypes & 47.4 & 35.0 & 21.0 & 36.5 \\
   $w^h(20)$ & 42.2 & 34.3 & \bf 39.8 & 38.3 \\
   $w^h(100)$ & 31.6  & \bf 44.1 & 31.9 & 37.0 \\
 ensemble($w^{hc}(20)$,$w^{hc}(100)$) (\bf continual) & 34.9  & 37.7  & 37.0  & 36.5  \\
ensemble($w^h(0)$,$w^h(20)$,$w^h(100)$)  & 40.6  & 39.7  & 34.8  & \bf 39.0\\
\midrule
\multicolumn{5}{l}{\bf Unsupervised initialisation, uniform} \\
\midrule
Prototypes & 40.3 & 38.1 & 29.9 & 37.1 \\
 
  Cosine classifier & \bf 48.9 & 27.0 & 13.2 & 32.0 \\
  Classifier + prototypes & 46.4 & 37.0 & 24.1 & 37.7 \\
   $w^h(20)$ & 43.0 & 35.2 & \bf 41.5 & 39.4 \\
   $w^h(100)$ & 33.5 & \bf 45.8 & 34.5 & 39.0 \\
 ensemble($w^{hc}(20)$,$w^{hc}(100)$) (\bf continual) & 38.9  & 38.9  & 37.7  & 38.7 \\
ensemble($w^h(0)$,$w^h(20)$,$w^h(100)$)  & 42.0  & 41.7  & 34.3  & \bf 40.2\\
\midrule

Disalign R101~\cite{disalign}  & 39.1& 42.0 & 29.1&  38.5  \\
\bottomrule
\end{tabular}
}
\end{table*}

\end{document}